\documentclass[letterpaper]{article} 
\usepackage[preprint]{aaai2027}  

\usepackage[hyphens]{url}  
\usepackage{graphicx} 
\usepackage{natbib}  
\usepackage{caption} 
\usepackage{algorithm}
\usepackage{algorithmic}

\usepackage{newfloat}
\usepackage{amsfonts}
\usepackage{amssymb}

\usepackage{amsmath}
\usepackage{listings}
\DeclareCaptionStyle{ruled}{labelfont=normalfont,labelsep=colon,strut=off} 
\floatstyle{ruled}
\newfloat{listing}{tb}{lst}{}
\floatname{listing}{Listing}

\usepackage{booktabs}

\title{PACE: Progressive Angular-to-Norm Contrastive Embedding}
\author{
    Yanping Li$^{1,3}$\equalcontrib, Wei Zhou$^{2,3}$\equalcontrib, Yawen Liu$^{3}$, Yibo Wang$^{3,4}$, Ke Zhu$^{3}$,\\ Guangda Huzhang$^{3}$, Qing-Guo Chen$^{3}$, Zhao Xu$^{3}$, Jun Zhang$^{1}$\corresponding, Wei Wei$^{2}$\corresponding \\
    $^1$The Hong Kong University of Science and Technology\\
    $^2$Huazhong University of Science and Technology\\
    $^3$Alibaba Group
    $^4$Nanjing University
}
\affiliations{
}

\begin{document}

\maketitle

\begin{abstract}
Multimodal embedding models encode heterogeneous inputs into a shared embedding
space, enabling efficient similarity computation across modalities and tasks.
Most existing methods optimize cosine-based contrastive objectives, which
promote stable training but restrict semantic compatibility to angular
geometry, precluding embedding norms from serving as an additional semantic
signal. However, directly optimizing the more expressive dot-product
similarity, which leverages both angular and norm information, underperforms
cosine-based training and exhibits unstable training dynamics. We attribute
this discrepancy to premature optimization-space expansion, manifested as
angular--norm entanglement and directional anisotropy in the representation space and further
compounded by full-parameter fine-tuning. In this paper, we propose
\textbf{PACE}, a two-stage framework that progressively expands both the
representation and trainable parameter spaces. Stage~I combines cosine-based
objective with low-rank adaptation to establish a reliable angular geometry
within constrained optimization spaces. Stage~II switches to dot-product
similarity and full-parameter fine-tuning, enabling embedding directions and
norms to jointly encode semantic information. We further introduce
\textbf{Focal Embedding Loss}, a confidence-adaptive objective that downweights
queries with high positive retrieval confidence while emphasizing ambiguous
queries with competitive negatives. Experiments across multiple backbone
scales and diverse multimodal embedding tasks consistently validate the
effectiveness of PACE.
\end{abstract}

\section{Introduction}

Multimodal embedding
models~\citep{li2026qwen3,shanbhogue2026gemini,lee2025gemini,zhang2025qwen3,radford2021clip,girdhar2023imagebind,zhu2024languagebind}
encode heterogeneous inputs---including text, images, and arbitrary
compositions thereof---into compact dense vectors in a shared embedding space,
such that semantically corresponding instances receive high similarity scores
regardless of their modalities. By providing a unified and modality-agnostic
representation interface, these models enable efficient similarity computation
across diverse input forms and support a broad spectrum of applications,
including cross-modal retrieval, classification, visual question answering,
and visual grounding~\citep{radford2021clip,jiang2024vlm2vec,lin2025mm,li2026token,li2026gear,xue2025vlmq}. Recent
research~\citep{jiang2024vlm2vec,lin2025mm,zhang2025gme,gu2026unime} has further shifted toward
universal multimodal embedding models that generalize across both input
modalities and task formulations within a single model.
Learning representations that are well aligned across modalities,
discriminative with respect to fine-grained semantics, and transferable across
tasks has therefore become a fundamental problem in multimodal representation
learning.

\begin{figure}[!t]
  \centering
  \includegraphics[width=\columnwidth]{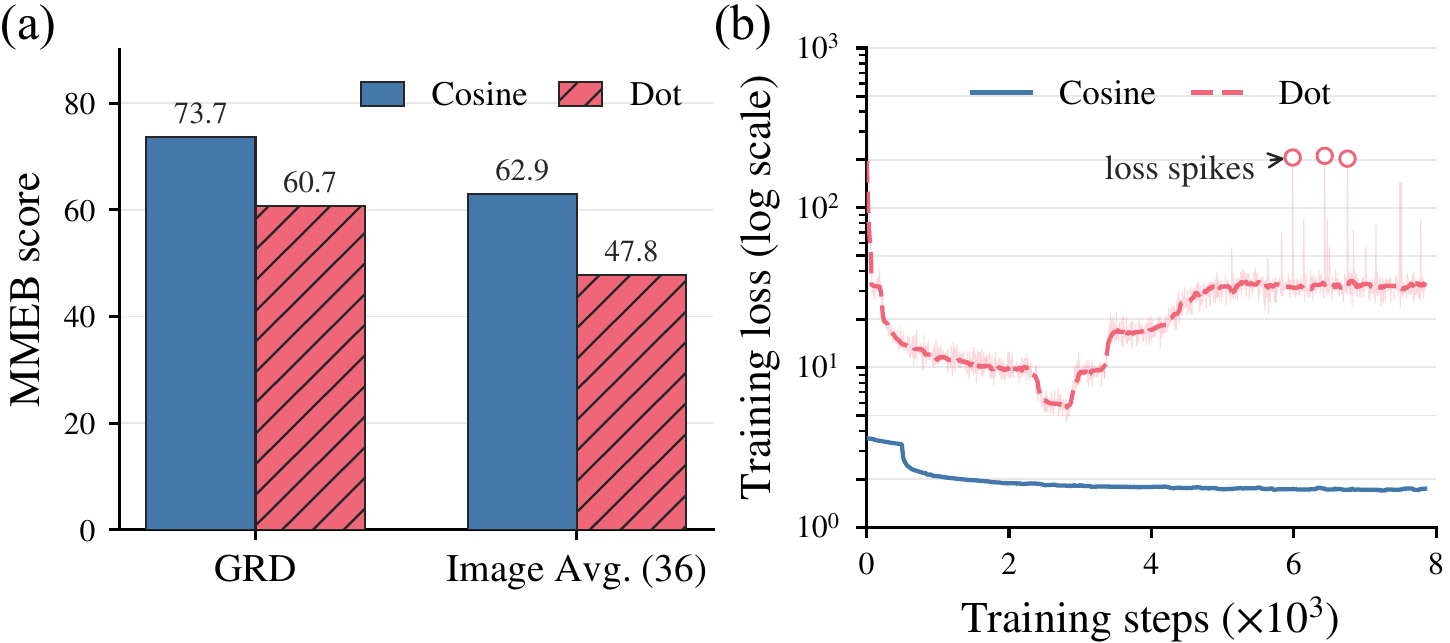}
  \caption{Comparison of direct dot-product and cosine-based optimization over
  a training epoch. (a) Direct dot-product optimization achieves lower embedding performance (MMEB score). (b) Its loss trajectory exhibits recurrent late-stage
  spikes, whereas cosine-based optimization remains comparatively stable.}
  \label{fig:direct-dot}
\end{figure}

Despite substantial advances in model
architectures~\citep{shanbhogue2026gemini,li2026qwen3} and multimodal data
curation~\citep{gadre2023datacomp,fang2024dfn,xu2024metaclip,zhang2025gme},
the dominant training paradigm for multimodal embedding models remains
geometrically restrictive. Most existing methods optimize an InfoNCE objective
using cosine similarity~\citep{radford2021clip,jiang2024vlm2vec,lin2025mm,zhu2026flash},
which computes similarity between $\ell_2$-normalized embeddings. By
eliminating radial degrees of freedom, normalization prevents norm variations
from rescaling contrastive logits and thereby simplifies optimization. However,
it restricts semantic compatibility to angular information, excluding embedding
norms as a potential relevance signal. This restriction is increasingly
consequential as multimodal large language models (MLLMs) are adapted into
universal multimodal embedders, since an angular-only similarity geometry may
underutilize their substantial representational capacity. Unnormalized inner
products have long served as relevance scores in retrieval and recommendation,
motivating extensive research on maximum inner-product search
(MIPS)~\citep{ram2012maximum,shrivastava2014asymmetric}. Recent evidence further
indicates that embedding norms can encode task-dependent relevance
information~\citep{feng2026magnitude}, motivating dot-product similarity as a
more expressive alternative that leverages both angular and norm information.
Nevertheless, our empirical results in Figure~\ref{fig:direct-dot} show that
directly optimizing dot-product similarity underperforms its cosine-based
counterpart and exhibits unstable training dynamics, revealing a persistent gap
between representational expressivity and optimization stability.

To understand the performance gap, preliminary experiments reveal two
challenges in direct dot-product optimization. First, jointly learning angular
alignment and norm scaling from the outset can entangle directions and norms,
causing them to encode redundant rather than complementary semantic
information. Cosine similarity avoids this interaction by removing radial
degrees of freedom. Second, dot-product training can induce directional
anisotropy, concentrating embeddings within a narrow cone and assigning
spuriously high similarity to semantically unrelated instances. Angular--norm
entanglement and directional anisotropy help explain its unexpected
underperformance and unstable training dynamics. Full-parameter fine-tuning
further compounds the difficulty by substantially enlarging the trainable
parameter space. Together, these findings motivate a coarse-to-fine strategy
that establishes a reliable angular geometry within constrained representation
and parameter spaces before progressively expanding both to exploit additional
capacity.

Based on this insight, we propose \textbf{P}rogressive
\textbf{A}ngular-to-Norm \textbf{C}ontrastive \textbf{E}mbedding
(\textit{PACE}), a two-stage framework for stable and expressive multimodal
embedding learning. PACE progressively expands both the representation and
trainable parameter spaces. Stage~I combines a cosine-based objective
with LoRA~\citep{hu2022lora}, restricting semantic organization to angular
geometry and parameter updates to a compact low-rank subspace. Initialized from
this constrained solution, Stage~II adopts dot-product similarity and
full-parameter fine-tuning, enabling embedding directions and norms to jointly
encode semantic information while exploiting the complete parameter space. A
complementary first-order gradient analysis formalizes the radial--angular
coupling induced by dot-product optimization and shows that, the progressive Stage~II initialization yields
locally weak cross-Jacobian interactions, providing a mechanism for mitigating
angular--norm entanglement.
Despite these benefits, multi-stage training may overemphasize already well-separated examples, increasing the risk of overfitting.
To mitigate this issue, we further
introduce \textbf{Focal Embedding Loss}, a confidence-adaptive objective
inspired by focal learning~\citep{lin2017focal}. It assigns each query a
difficulty-aware weight based on its positive retrieval confidence,
downweighting examples whose positives are already well separated from
negatives while emphasizing ambiguous examples with competitive negatives.
This mechanism suppresses redundant updates on saturated examples and directs
the expanded optimization capacity toward unresolved semantic distinctions.
Extensive experiments across multiple model scales and a diverse suite of multimodal embedding benchmarks demonstrate that PACE consistently outperforms strong baselines and validating the effectiveness of each proposed component.
Our main contributions are summarized as follows:

\begin{itemize}
    \item To our knowledge, we provide the first systematic diagnosis that
    attributes the unexpected underperformance and unstable training dynamics
    of direct dot-product optimization to angular--norm entanglement and
    directional anisotropy, supported by controlled experiments and a
    complementary theoretical analysis.
    \item We propose PACE, a progressive training framework that combines a
    cosine-to-dot-product curriculum with a LoRA-to-full-fine-tuning schedule,
    progressively expanding the representation and trainable parameter spaces
    for stable and expressive multimodal embedding learning.
    \item We further introduce Focal Embedding Loss, a confidence-adaptive objective
    that downweights well-separated examples and emphasizes ambiguous examples
    with competitive negatives. Experiments across multiple backbone scales and
    diverse multimodal embedding tasks demonstrate the effectiveness of the
    overall framework.
\end{itemize}

\begin{figure*}[!t]
  \centering
  \includegraphics[width=\textwidth]{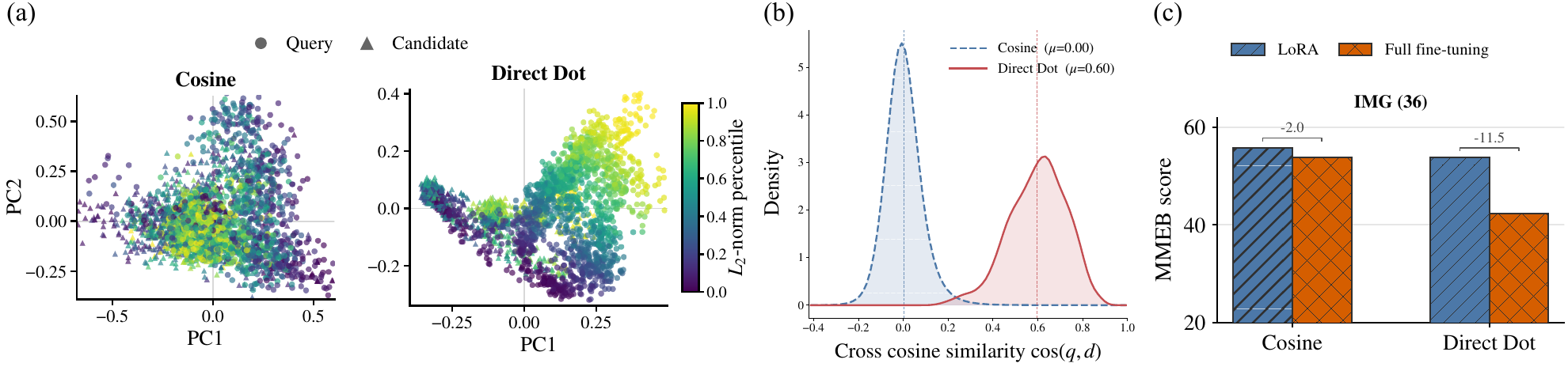}
  \caption{Diagnostic analyses of direct dot-product training. (a) PCA
  projections of $\ell_2$-normalized embeddings, colored by their original
  norm percentiles. Direct dot-product training produces a norm-stratified
  directional geometry absent under cosine-based training. (b) Cosine-similarity
  distributions between randomly paired queries and candidates, showing
  substantially greater directional concentration under direct dot-product
  training. (c) MMEB image-task performance under LoRA and full-parameter
  fine-tuning, where direct dot-product training exhibits a markedly larger
  degradation.}
  \label{fig:motivation}
\end{figure*}

\section{Related Work}

Existing work improves embedding models through four complementary directions:
model development, query augmentation, data-side techniques, and similarity
design.

Multimodal embedding learning has progressed from vision--language alignment
to universal representation learning. CLIP~\citep{radford2021clip} establishes
the image--text dual-encoder paradigm, while
ImageBind~\citep{girdhar2023imagebind} and
LanguageBind~\citep{zhu2024languagebind} extend shared-space learning to
additional modalities. More recently, MLLMs have been adapted into universal
embedders through instruction-aware training and broader task and modality
coverage~\citep{jiang2024vlm2vec,lin2025mm,zhang2025gme}. Native
multimodal embedding models further scale model capacity and modality
coverage~\citep{shanbhogue2026gemini,li2026qwen3}.

Query augmentation improves retrieval by enriching the input representation.
Query2doc~\citep{wang2023query2doc} and HyDE~\citep{gao2023hyde} generate
pseudo-documents, BRIGHT~\citep{su2025bright} studies explicit reasoning for
retrieval, and ExpandR~\citep{yao2025expandr} aligns generated expansions with
the retriever. These methods generally require additional LLM inference for
each query.

Data-side techniques encompass data curation and difficulty-aware supervision.
Existing work improves training data through standardized selection, learned
filtering, metadata-guided balancing, and task-diverse
synthesis~\citep{gadre2023datacomp,fang2024dfn,xu2024metaclip,zhang2025gme}.
For difficulty-aware training, MM-Embed~\citep{lin2025mm} mines
modality-aware hard negatives, while
Focal-InfoNCE~\citep{hou2023focalinfonce} applies pair-level focal modulation.
Query-level reweighting based on the positive retrieval probability over the
candidate set remains underexplored. Our Focal Embedding Loss addresses this
gap by downweighting well-separated queries and emphasizing those with
competitive negatives.

Most multimodal embedders use cosine similarity over normalized
representations, excluding embedding norms from similarity computation.
Unnormalized inner products have long served as retrieval and recommendation
scores, motivating maximum inner-product
search~\citep{ram2012maximum,shrivastava2014asymmetric}. Feng and
Watanabe~\citep{feng2026magnitude} further show that query and candidate norms
play distinct, task-dependent roles in optimization and inference. However,
reliably learning a multimodal embedding geometry that jointly exploits angular
and norm information remains underexplored.

\section{Methodology}
\label{sec:methodology}

\begin{figure*}[!t]
  \centering
  \includegraphics[width=0.8\textwidth]{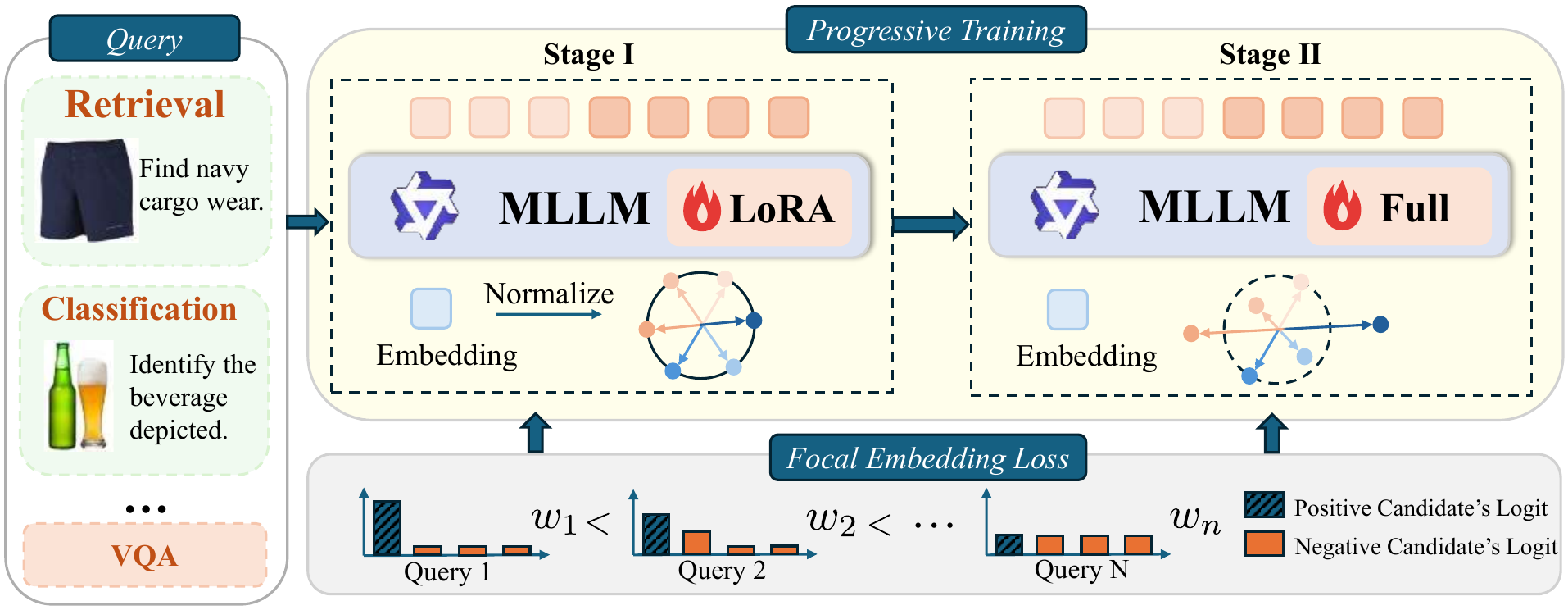}
  \caption{Overview of \textbf{PACE}. Stage~I learns a stable angular
  geometry using cosine similarity and LoRA within constrained representation
  and parameter spaces. Stage~II expands to dot-product similarity and
  full-parameter fine-tuning, while Focal Embedding Loss emphasizes ambiguous
  queries with competitive negatives.}
  \label{fig:pipeline}
\end{figure*}

\subsection{Problem Formulation}
\label{sec:problem-formulation}
Given a multimodal query $x_i^q$ and a candidate set
$\mathcal{C}_i=\{x_i^+,x_{i,1}^-,\ldots,x_{i,K}^-\}$, both queries and
candidates may consist of text, images, or interleaved image--text sequences.
We employ a shared MLLM-based encoder $f_\theta$ to map each heterogeneous input
into a $d$-dimensional dense vector in a shared embedding space:
\[
\mathbf{q}_i=f_\theta(x_i^q),\qquad
\mathbf{c}=f_\theta(x),\quad x\in\mathcal{C}_i.
\]
Candidate relevance is measured by a similarity function
$s(\mathbf{q}_i,\mathbf{c})$, according to which the candidate set is ranked.
The learning objective is to assign the positive candidate $x_i^+$ a higher
similarity score than all negative candidates, thereby supporting
modality-agnostic retrieval across heterogeneous input compositions.

\subsection{Motivation}
\label{sec:motivation}

To diagnose why direct dot-product optimization exhibits unstable training
dynamics, we examine whether it exploits angular and norm information
complementarily. Using cosine-based training as a controlled reference under
matched backbones, data, and training settings, we analyze 5,000 randomly
sampled training instances and assess whether full-parameter fine-tuning
amplifies the optimization difficulty.

\noindent\textbf{Angular--Norm Entanglement.}
After $\ell_2$-normalizing the learned embeddings, we apply PCA to visualize
their directional structure and color each point by its original norm. As shown
in Figure~\ref{fig:motivation}(a), embeddings with substantially different
norms are intermixed under cosine-based training, indicating little visible
dependence between directions and norms. Direct dot-product training instead
produces a norm-stratified structure in which embeddings with similar norms
occupy nearby projected regions. This qualitative pattern suggests that direct
optimization entangles angular and norm information rather than exploiting
them complementarily.

\noindent\textbf{Directional Anisotropy.}
We next examine cosine similarities between randomly paired queries and
candidates. Figure~\ref{fig:motivation}(b) shows that cosine-based training
yields lower random-pair similarities and a more dispersed directional
geometry. Direct dot-product training shifts the distribution toward higher
similarity values, indicating that embedding directions concentrate within a
narrower region of the unit hypersphere. Such directional concentration can
increase spurious similarities between semantically unrelated instances and
potentially exacerbate hubness~\cite{radovanovic2010hubs,angiulli2018behavior}.

\noindent\textbf{Effect of Parameter-Space Expansion.}
Finally, Figure~\ref{fig:motivation}(c) compares cosine- and dot-product
training under LoRA and full-parameter fine-tuning. Full-parameter fine-tuning
underperforms LoRA for both similarity functions, with a substantially larger
degradation under dot-product similarity. This interaction suggests that
simultaneously enlarging the representation and trainable parameter spaces
compounds early-stage optimization difficulty.

Overall, direct dot-product optimization does not necessarily realize its
greater representational capacity. These findings motivate a coarse-to-fine
strategy that first establishes a reliable angular geometry within a compact
parameter subspace and subsequently introduces radial degrees of freedom and
full-parameter updates.

\subsection{PACE}
As illustrated in Figure~\ref{fig:pipeline}, PACE is a two-stage framework
that progressively expands both the representation and trainable parameter
spaces. Stage~I restricts optimization to angular similarity and a compact
low-rank parameter subspace, providing a stable initialization with a reliable
semantic geometry. Stage~II subsequently introduces radial representational
freedom and full-parameter optimization, enabling the model to exploit the
increased capacity of both spaces. Focal Embedding Loss is applied in both
stages. Using a consistent cosine-normalized confidence function across both
stages mitigates cross-stage shifts in the focal-weight distribution and
prevents embedding-norm scaling from distorting query difficulty.

\noindent \textbf{Stage~I: Constrained Optimization Space.}
In the first stage, we restrict the similarity function to angular information
by $\ell_2$-normalizing query and candidate embeddings:
\begin{equation}
s^{(1)}(\mathbf{q},\mathbf{c})
=
s_{\mathrm{cos}}(\mathbf{q},\mathbf{c})
=
\frac{\mathbf{q}^{\top}\mathbf{c}}
{\|\mathbf{q}\|_2\|\mathbf{c}\|_2}.
\label{eq:stage1-similarity}
\end{equation}
The resulting scale invariance removes radial degrees of freedom from the
contrastive objective, allowing the model to first establish a well-structured
angular geometry. We simultaneously restrict parameter updates using
LoRA~\citep{hu2022lora}. For a pretrained weight matrix $\mathbf{W}_0$, LoRA
parameterizes its update as $\Delta\mathbf{W}=\mathbf{B}\mathbf{A}$, where
$\mathrm{rank}(\Delta\mathbf{W})\leq r$ and $r$ is substantially smaller than
the dimensions of $\mathbf{W}_0$. Optimizing only these low-rank updates
constrains training to a compact parameter subspace, reducing early-stage
optimization difficulty and providing a stable, semantically meaningful
initialization for the subsequent stage.

\noindent \textbf{Stage~II: Expanded Optimization Space.}
The second stage initializes from the Stage~I solution and expands both
optimization spaces. In the representation space, we remove $\ell_2$
normalization and adopt dot-product similarity:
\begin{equation}
s^{(2)}(\mathbf{q},\mathbf{c})
=
s_{\mathrm{dot}}(\mathbf{q},\mathbf{c})
=
\mathbf{q}^{\top}\mathbf{c}.
\label{eq:stage2-similarity}
\end{equation}
This transition releases the radial degrees of freedom, allowing embedding
directions and norms to jointly encode semantic information. In the parameter
space, we merge the learned LoRA updates into the pretrained backbone and
perform full-parameter fine-tuning. Starting from the constrained Stage~I
solution allows the model to exploit the greater representational and parameter
capacity of Stage~II without directly optimizing both expanded spaces from the
beginning.

\noindent \textbf{Focal Embedding Loss.}
Although progressive expansion improves representational capacity, both stages
may overemphasize queries whose positive candidates are already well separated
from the negatives. We introduce Focal Embedding Loss, which
adaptively reweights each query according to its current retrieval difficulty.
For Stage~$T\in\{\mathrm{I},\mathrm{II}\}$, the positive probability used by the contrastive
objective is
\begin{equation}
\pi_i^{(t)}
=
\frac{
\exp\left(s^{(t)}(\mathbf{q}_i,\mathbf{c}_i^+)/\tau_t\right)
}{
\sum_{\mathbf{c}\in\mathcal{C}_i}
\exp\left(s^{(t)}(\mathbf{q}_i,\mathbf{c})/\tau_t\right)
},
\label{eq:stage-positive-probability}
\end{equation}
where $s^{(t)}$, $t\in\{1,2\}$, denotes the similarity employed in Stage~$T$; specifically, $s^{(1)}$ and $s^{(2)}$ are the cosine and dot-product similarities used in Stages~I and~II, respectively.
And $\tau_t$ is the corresponding contrastive temperature. The
per-query contrastive loss is $\ell_i^{(t)}=-\log\pi_i^{(t)}$.

To obtain a norm-invariant estimate of query difficulty, we compute the focal
confidence using cosine-normalized logits in both stages:
\begin{equation}
p_i^{(t)}
=
\frac{
\exp\left(
s_{\mathrm{cos}}(\mathbf{q}_i,\mathbf{c}_i^+)/\tau_{\mathrm{w}}
\right)
}{
\sum_{\mathbf{c}\in\mathcal{C}_i}
\exp\left(
s_{\mathrm{cos}}(\mathbf{q}_i,\mathbf{c})/\tau_{\mathrm{w}}
\right)
},
\label{eq:focal-confidence}
\end{equation}
where $\tau_{\mathrm{w}}$ is the weighting temperature. Decoupling the
weighting confidence from the stage-specific training similarity prevents
embedding-norm scaling in Stage~II from artificially inflating the estimated
confidence and provides a consistent difficulty measure across stages.

We first compute a detached focal coefficient
\begin{equation}
a_i^{(t)}
=
\mathrm{sg}\!\left[
\left(1-p_i^{(t)}\right)^\gamma
\right],
\label{eq:focal-weight}
\end{equation}
where $\mathrm{sg}[\cdot]$ denotes stop-gradient and $\gamma$ determines the
strength of difficulty modulation. We normalize these coefficients to have
unit mean over a minibatch of $N$ queries:
\begin{equation}
\widetilde{w}_i^{(t)}
=
\frac{a_i^{(t)}}{\frac{1}{N}\sum_{k=1}^{N}a_k^{(t)}}.
\label{eq:normalized-focal-weight}
\end{equation}
The focal-weighted objective for Stage~$t$ is then
\begin{equation}
\mathcal{L}_{\mathrm{PACE}}^{(t)}
=
\frac{1}{N}\sum_{i=1}^{N}\widetilde{w}_i^{(t)}\ell_i^{(t)}
=
-\frac{\sum_{i=1}^{N}a_i^{(t)}\log\pi_i^{(t)}}
{\sum_{i=1}^{N}a_i^{(t)}}.
\label{eq:focal-embedding-loss}
\end{equation}
Queries whose positives are angularly well separated from the negatives yield
high confidence and receive smaller weights, whereas ambiguous queries with
competitive negatives receive larger weights. This mechanism suppresses
redundant updates on saturated examples and directs optimization toward
unresolved semantic distinctions in both training stages.

\subsection{Theoretical Analysis}
\label{sec:theoretical-analysis}
\noindent\textbf{Radial--Angular Gradient Decomposition.}
We analyze the first-order embedding dynamics of the unweighted InfoNCE
objectives in both stages. Let $\mathbf{q}=r\widehat{\mathbf{q}}$, where
$\|\widehat{\mathbf{q}}\|_2=1$, and define
$\mathbf{P}_{\widehat{\mathbf{q}}}^{\perp}
=\mathbf{I}-\widehat{\mathbf{q}}\widehat{\mathbf{q}}^{\top}$.
Differentiating the cosine objective through $\ell_2$ normalization gives
\begin{equation}
\nabla_{\mathbf{q}}\mathcal{L}_{\mathrm{cos}}
=
\frac{1}{r}\mathbf{P}_{\widehat{\mathbf{q}}}^{\perp}
\nabla_{\widehat{\mathbf{q}}}\mathcal{L}_{\mathrm{cos}},
\qquad
\widehat{\mathbf{q}}^{\top}
\nabla_{\mathbf{q}}\mathcal{L}_{\mathrm{cos}}=0.
\label{eq:cos-gradient}
\end{equation}
Hence, cosine-based training has no first-order radial embedding gradient.
Assume candidates share a common norm and write
$\mathbf{c}_j=R\widehat{\mathbf{c}}_j$ and
$u_j=\widehat{\mathbf{q}}^{\top}\widehat{\mathbf{c}}_j$. For dot-product InfoNCE, let
$p_j^{(2)}=\exp(rRu_j/\tau_2)/
\sum_k\exp(rRu_k/\tau_2)$. Its gradient decomposes as
\begin{equation}
\begin{aligned}
\nabla_{\mathbf{q}}\mathcal{L}_{\mathrm{dot}}
&=a_d\widehat{\mathbf{q}}+\mathbf{b}_d,\\
a_d
&=\frac{R}{\tau_2}\left(\mathbb{E}_{p^{(2)}}[u]-u_+\right),\\
\mathbf{b}_d
&=\frac{R}{\tau_2}\mathbf{P}_{\widehat{\mathbf{q}}}^{\perp}
\left(\sum_jp_j^{(2)}\widehat{\mathbf{c}}_j
-\widehat{\mathbf{c}}_+\right).
\end{aligned}
\label{eq:dot-gradient}
\end{equation}
Here, $a_d$ and $\mathbf{b}_d$ govern radial and angular evolution,
respectively; because $p^{(2)}$ depends jointly on $r$ and $u$, the two
dynamics are coupled.

\noindent\textbf{Proposition 1 (Confidence-Dependent Dynamic Decoupling).}
Under the common-norm setting above, consider the Stage~II initialization
$\mathbf{q}_0=r_0\widehat{\mathbf{q}}$ and let
$\beta=r_0R/\tau_2$. If its positive posterior satisfies
$p_+^{(2)}\geq 1-\rho$, then
  \begin{equation}
  \left\|\frac{\partial\mathbf{b}_d}{\partial r}\right\|_2
  \leq \frac{4R^2}{\tau_2^2}\rho,\quad
  \left\|D_{\widehat{\mathbf{q}}}a_d\right\|_{\mathrm{op}}
  \leq \frac{R}{\tau_2}(4\beta+2)\rho.
  \label{eq:cross-jacobian-bound}
  \end{equation}
where $D_{\widehat{\mathbf{q}}}a_d[\mathbf{h}]$ denotes the directional
derivative of the radial coefficient $a_d$ with respect to
$\widehat{\mathbf{q}}$ along a tangent perturbation $\mathbf{h}$, and
$\|D_{\widehat{\mathbf{q}}}a_d\|_{\mathrm{op}}
=\sup_{\mathbf{h}\perp\widehat{\mathbf{q}},\,\|\mathbf{h}\|_2=1}
|D_{\widehat{\mathbf{q}}}a_d[\mathbf{h}]|$.
These off-diagonal blocks characterize the local coupling between radial and
angular dynamics. For bounded $\beta$ and $R/\tau_2$, both cross-Jacobian
norms vanish as $\rho\rightarrow0$, yielding local first-order decoupling at
the Stage~II transition. This result is consistent with the reduced
angular--norm dependence observed empirically but does not imply global
statistical independence. A complete derivation is provided in the supplementary material.

\section{Experiment}
\label{sec:experiment}

\begin{table*}[t]
    \centering

    \small
    \setlength{\tabcolsep}{4.2pt}

    \begin{tabular}{l c c c c c c c c}
        \toprule
        \textbf{Models}
        & \textbf{\#Params}
        & \multicolumn{4}{c}{\textbf{Per Meta-Task Score}}
        & \multicolumn{3}{c}{\textbf{Average Score}} \\
        \cmidrule(lr){3-6}
        \cmidrule(lr){7-9}
        &
        & \textbf{Classification}
        & \textbf{VQA}
        & \textbf{Retrieval}
        & \textbf{Grounding}
        & \textbf{IND}
        & \textbf{OOD}
        & \textbf{Overall} \\
        \midrule

        \textbf{\# of Datasets}
        & -- & 10 & 10 & 12 & 4 & 20 & 16 & 36 \\
        \midrule

        \multicolumn{9}{c}{\textbf{Zero-shot on MMEB}} \\
        \midrule
        \addlinespace[0.5pt]

        CLIP (ViT-L)
        & 0.4B & 42.8 & 9.1 & 53.0 & 51.8 & 37.1 & 38.7 & 37.8 \\
        OpenCLIP (ViT-L)
        & 0.4B & 47.8 & 10.9 & 52.3 & 53.3 & 39.3 & 40.2 & 39.7 \\
        MagicLens (ViT-L)
        & 0.4B & 38.8 & 8.3 & 35.4 & 26.0 & 31.0 & 23.7 & 27.1 \\
        SigLIP (So/14)
        & 0.9B & 40.3 & 8.4 & 31.6 & 59.5 & 32.3 & 38.0 & 35.0 \\
        BLIP-2 (ViT-L)
        & 1.2B & 27.0 & 4.2 & 33.9 & 47.0 & 25.3 & 25.1 & 28.0 \\
        CLIP (ViT-BigG/14)
        & 2.5B & 52.3 & 14.0 & 50.5 & 60.3 & 38.9 & 45.8 & 44.3 \\
        EVA-CLIP
        & 8B & 56.0 & 10.4 & 49.2 & 58.9 & 38.1 & 45.6 & 43.7 \\
        E5-V (Phi-3.5-V)
        & 4.2B & 39.1 & 9.6 & 38.0 & 57.6 & 33.1 & 31.9 & 36.1 \\
        E5-V (LLaVA-1.6)
        & 7B & 39.7 & 10.8 & 39.4 & 60.2 & 34.2 & 33.4 & 37.5 \\

        \midrule
        \multicolumn{9}{c}{\textbf{Fine-tuning on MMEB}} \\
        \midrule
        \addlinespace[0.5pt]

        VLM2Vec (Qwen3.5-0.8b)
        & 0.8B & 51.0 & 56.8 & 53.1 & 73.2 & 61.1 & 49.2 & 55.8 \\
        VLM2Vec (Qwen3.5-2b)
        & 2B & 61.4 & 62.0 & 61.4 & 73.7 & 67.2 & 57.6 & 62.9 \\
        UniME-V2 (Qwen3.5-0.8b)
        & 0.8B & 51.7 & 59.0 & 53.8 & 75.0 & 60.9 & 52.2 & 57.0 \\
        UniME-V2 (Qwen3.5-2b)
        & 2B & 60.7 & 64.1 & 63.8 & 80.3 & 67.0 & 62.1 & 64.9 \\

        \addlinespace[0.5pt]
        \midrule
        \addlinespace[0.5pt]

        \textbf{PACE (Qwen3.5-0.8b)}
        & 0.8B & 52.3~(+0.6) & 60.4~(+1.4) & 57.4~(+3.6) & 77.2~(+2.2) & 63.8~(+2.7) & 53.0~(+0.8) & 59.0~(+2.0) \\
        \textbf{PACE (Qwen3.5-2b)}
        & 2B & \textbf{63.6}~(+2.2) & \textbf{66.0}~(+1.9) & \textbf{66.5}~(+2.7) & \textbf{81.8}~(+1.5) & \textbf{71.2}~(+4.0) & \textbf{62.4}~(+0.3) & \textbf{67.3}~(+2.4) \\

        \bottomrule
    \end{tabular}
    \caption{%
        Results on the MMEB benchmark.
        IND:~in-distribution, OOD:~out-of-distribution.
        Scores are average Precision@1 (\%).
        The best are highlighted in
        \textbf{bold}. $\Delta$ values in parentheses are computed against the highest-scoring baseline of the same model. Detailed results are in the supplementary materials.
    }
    \label{tab:mmeb_main_results}
\end{table*}
\subsection{Implementation}
\label{sec:implementation}

We train PACE using two 
MLLMs,
Qwen3.5-0.8B-Base and Qwen3.5-2B-Base~\citep{qwen35}. Training consists of two
stages. In the first stage, we implement LoRA~(rank=32) and adopt Focal Embedding Loss with global weight normalization,
setting the focal parameter $\gamma$ to 2.0 and using a learning rate of $2\times10^{-5}$.
In the second stage, we initialize the model from the step-2,000 checkpoint obtained in the first stage and perform full-parameter language-model fine-tuning with Focal Embedding Loss, where the similarity metric is calculated using the dot product.
The learning rate is reduced to $2\times10^{-6}$.
In both stages, we use 5 hard negatives per query. 
Detailed 
hyperparameters and hardware
configurations are provided in the supplementary material.

\subsection{Datasets and Evaluation}
\label{sec:datasets and evaluation}

\subsubsection{Training Data.}
\label{sec:training data.}

We use the mmE5-Synthetic dataset~\citep{chen2025mme5} as our training data. After preprocessing, the resulting corpus comprises approximately 2.0 million training instances constructed under an MMEB-oriented design for general-purpose multimodal embedding learning. It provides task-relevant supervision across MMEB-style task formats and modality combinations, aligning with the MMEB benchmark, which consists of 20 in-distribution and 16 out-of-distribution datasets spanning classification, visual question answering, multimodal retrieval, and visual grounding.

\subsubsection{Evaluation.}
\label{sec:evaluation}
We evaluate PACE on both the in-distribution and out-of-distribution splits of MMEB~\citep{jiang2024vlm2vec}, comprising 20 and 16 test sets, respectively, to assess its general-purpose multimodal embedding capabilities across diverse tasks. Following the standard MMEB evaluation protocol, we report Precision@1, defined as the proportion of queries for which the correct candidate is ranked first. Unlike conventional cosine-similarity evaluation, PACE ranks candidates using the dot product between unnormalized query and candidate embeddings, consistent with the dot-product objective employed during the second training stage.

\subsection{Baselines}

We compare PACE against both zero-shot and fine-tuned multimodal embedding
models. The zero-shot baselines include general-purpose vision--language
encoders, namely CLIP~\citep{radford2021clip},
OpenCLIP~\citep{cherti2023openclip},
MagicLens~\citep{zhang2024magiclens},
SigLIP~\citep{zhai2023siglip}, BLIP-2~\citep{li2023blip2}, and
EVA-CLIP~\citep{sun2023evaclip}, as well as MLLM-based embedders, including
E5-V~\citep{jiang2024e5} and pretrained Qwen3.5
backbones
without embedding-specific fine-tuning. For
fine-tuned comparisons, we consider VLM2Vec~\citep{jiang2024vlm2vec}, which
performs instruction-guided contrastive training with standard InfoNCE loss across diverse multimodal
embedding tasks. UniME-V2~\citep{gu2026unime} further employs MLLM-derived
semantic matching scores to improve contrastive supervision.

\section{Main Results}
\label{sec:main-results}

Table~\ref{tab:mmeb_main_results} presents the quantitative results on MMEB, where all methods are trained under identical
hyperparameter settings for fair comparison.
PACE achieves overall scores of 59.0 and 67.3 with the 0.8B and 2B backbones,
respectively; the 0.8B model alone exceeds the strongest zero-shot baseline
by 14.7 despite using fewer parameters. Against the strongest scale-matched
fine-tuned baselines, PACE yields overall margins of 2.0 and 2.4, with
per-category improvements of 0.6--3.6 (0.8B) and 1.5--2.7 (2B) across
classification, VQA, retrieval, and grounding. Positive margins on both IND
and OOD splits further confirm robustness across evaluation distributions.
These consistent gains support the central premise of PACE: establishing a
reliable angular geometry before progressively expanding the representation
and parameter spaces allows MLLM backbones to realize their representational
capacity without sacrificing optimization stability.

\section{Geometric Analysis}
\label{sec:geometric-analysis}

\begin{figure}[!t]
  \centering
  \includegraphics[width=\columnwidth]{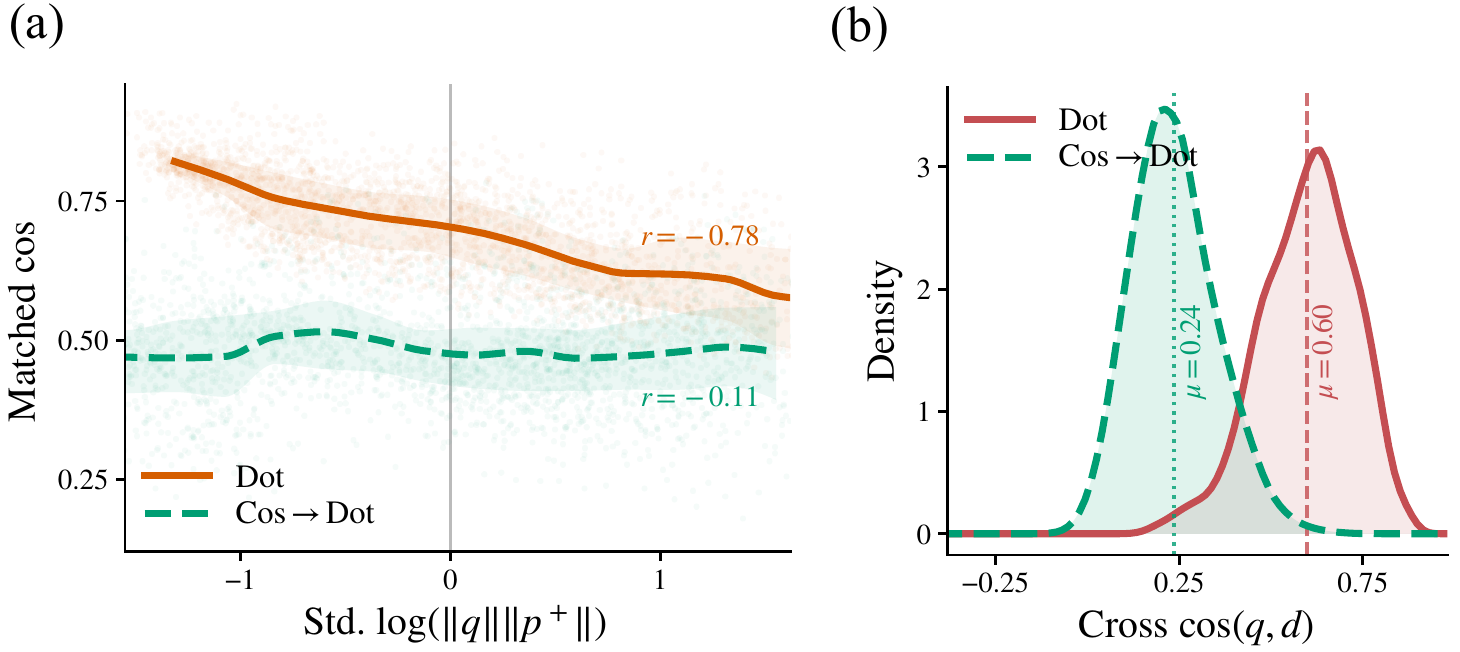}
  \caption{Angular-geometry diagnostics of direct dot-product training vs. PACE.
(a)~PACE weakens the negative correlation between cosine similarity and the
standardized log norm product. (b)~PACE lowers the mean cross-cosine
similarity, indicating a more angularly dispersed embedding geometry.}
  \label{fig:positive-coupling_and_cross_cosine}
\end{figure}

To assess whether PACE alleviates the representation-space distortions
caused by direct dot-product training, we randomly analyze a subset of training data, focusing on the angular-norm coupling and directional anisotropy. 
Additional visual embedding geometry analysis are provided in the supplementary material.

\noindent\textbf{Reduced Angular--Norm Coupling.}
Figure~\ref{fig:positive-coupling_and_cross_cosine}(a) plots the standardized
$\log(\lVert\mathbf{q}\rVert_2\lVert\mathbf{c}^{+}\rVert_2)$ against cosine
similarity for matched query--positive pairs. Direct dot-product training
yields a 
negative Pearson correlation ($r=-0.78$), indicating
strong angular--norm coupling, whereas PACE reduces it to $r=-0.11$. This
near-zero correlation implies little mutual information between the two
factors: instead of encoding redundant information as in the coupled regime,
norm and direction carry complementary signals, enlarging the effective
representational capacity of the embedding space.

\noindent\textbf{Directional Dispersion.}
Figure~\ref{fig:positive-coupling_and_cross_cosine}(b) compares the cosine-similarity
distributions of randomly paired queries and candidates. PACE yields a lower
mean similarity than direct dot-product training, consistent with reduced
global directional concentration and a more angularly dispersed embedding
geometry. Together, these results demonstrate that PACE mitigates both
angular--norm coupling and directional anisotropy.

\section{Ablation Studies}
\label{sec:ablation}

\noindent\textbf{Component Contributions.}
Table~\ref{tab:ablation_training_strategy} evaluates the contribution of each
PACE component. Removing Focal Embedding Loss decreases the overall score from
59.0 to 57.7. Disabling representation-space expansion by retaining cosine
similarity in Stage~II further reduces the score to 57.2, while retaining LoRA
instead of transitioning to full-parameter fine-tuning yields the largest
degradation, to 54.1. These results demonstrate that difficulty-aware
weighting, cosine-to-dot representation-space expansion, and LoRA-to-full
parameter-space expansion each contribute to 
PACE.

\begin{table}[t]
    \centering
    \setlength{\tabcolsep}{5.0pt}
    \small

    \begin{tabular}{l c c c c c}
        \toprule
        \textbf{Method}
        & \textbf{ClS}
        & \textbf{VQA}
        & \textbf{RET}
        & \textbf{GRD}
        & \textbf{Overall} \\
        \midrule

        \textbf{PACE~(Ours)}
        & 52.3 & 60.4 & 57.4 & 77.2 & 59.0\\

        PACE w/o Focal loss
        & 52.3 & 59.7 & 55.0 & 74.1 & 57.7\\

        PACE w/o Dot
        & 52.1 & 58.2 & 54.3 & 76.0 & 57.2\\

        PACE w/ S2 LoRA
        & 49.8 & 56.4 & 49.6 & 72.3 & 54.1\\

        \bottomrule
    \end{tabular}
    \caption{
        Ablation study examining effects of training objectives and second-stage tuning
        strategy on MMEB. 
        Abbreviations: CLS, Classification; RET, Retrieval; GRD, Grounding. S2 denotes Stage~II.
    }
    \label{tab:ablation_training_strategy}
\end{table}

\begin{figure}[!t]
  \centering
  \includegraphics[width=\columnwidth]{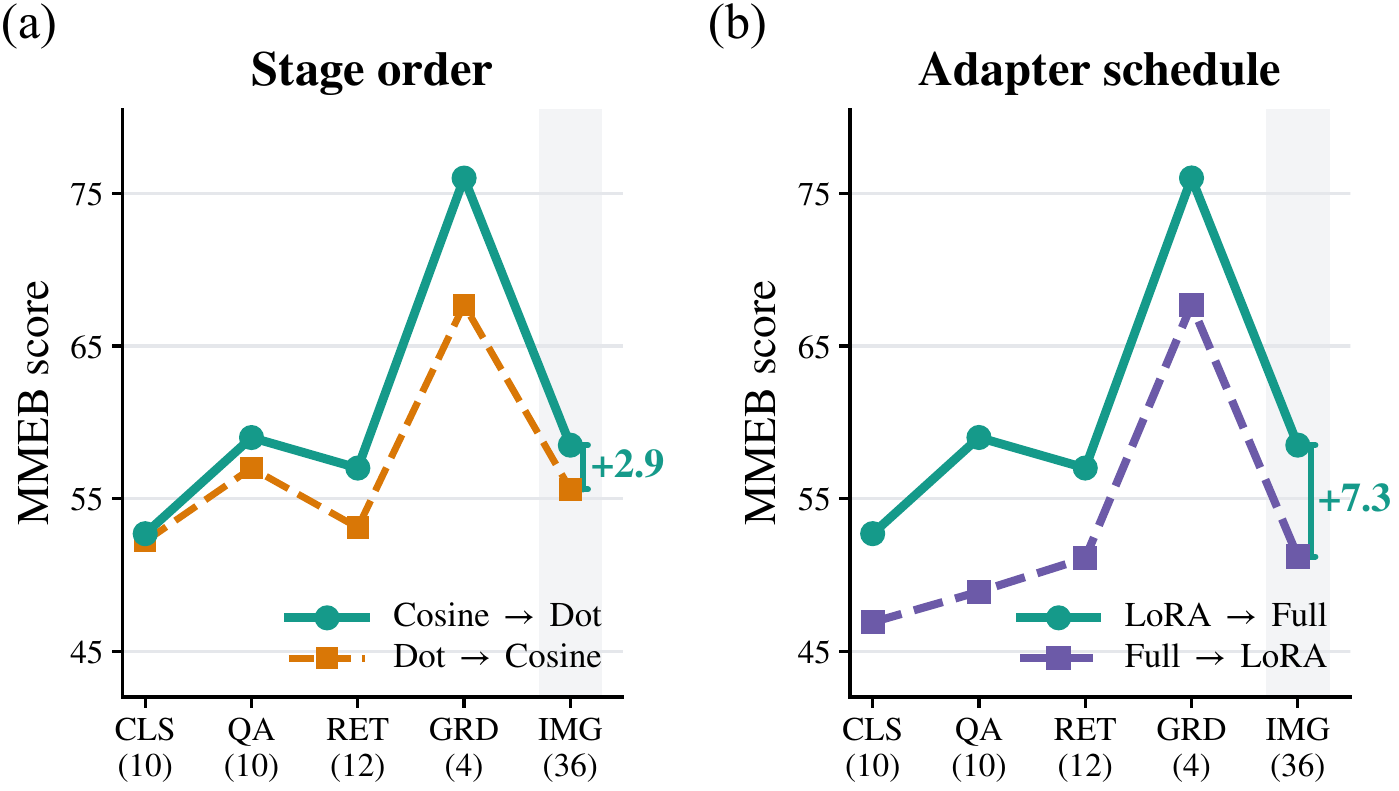}
  \caption{Effect of progressive expansion order on MMEB. Cosine-to-dot and
  LoRA-to-full schedules outperform their reversed counterparts by 2.9 and 7.3
  overall, respectively.}
  \label{fig:protocol_and_schedule}
\end{figure}

\noindent\textbf{Order of Optimization-Space Expansion.}
To determine whether the improvement arises from the progressive expansion
order rather than two-stage training alone,
Figure~\ref{fig:protocol_and_schedule} compares PACE with the corresponding
reversed schedules. Cosine-to-dot training outperforms dot-to-cosine training
by 2.9 overall, demonstrating the benefit of establishing an angular
semantic geometry before introducing radial degrees of freedom. Similarly,
LoRA-to-full fine-tuning exceeds the reversed full-to-LoRA schedule by 7.3. This result supports our hypothesis that early optimization within a
compact parameter subspace provides a more reliable update direction for
subsequent full-parameter optimization.

\begin{figure}[!t]
  \centering
  \includegraphics[width=\columnwidth]{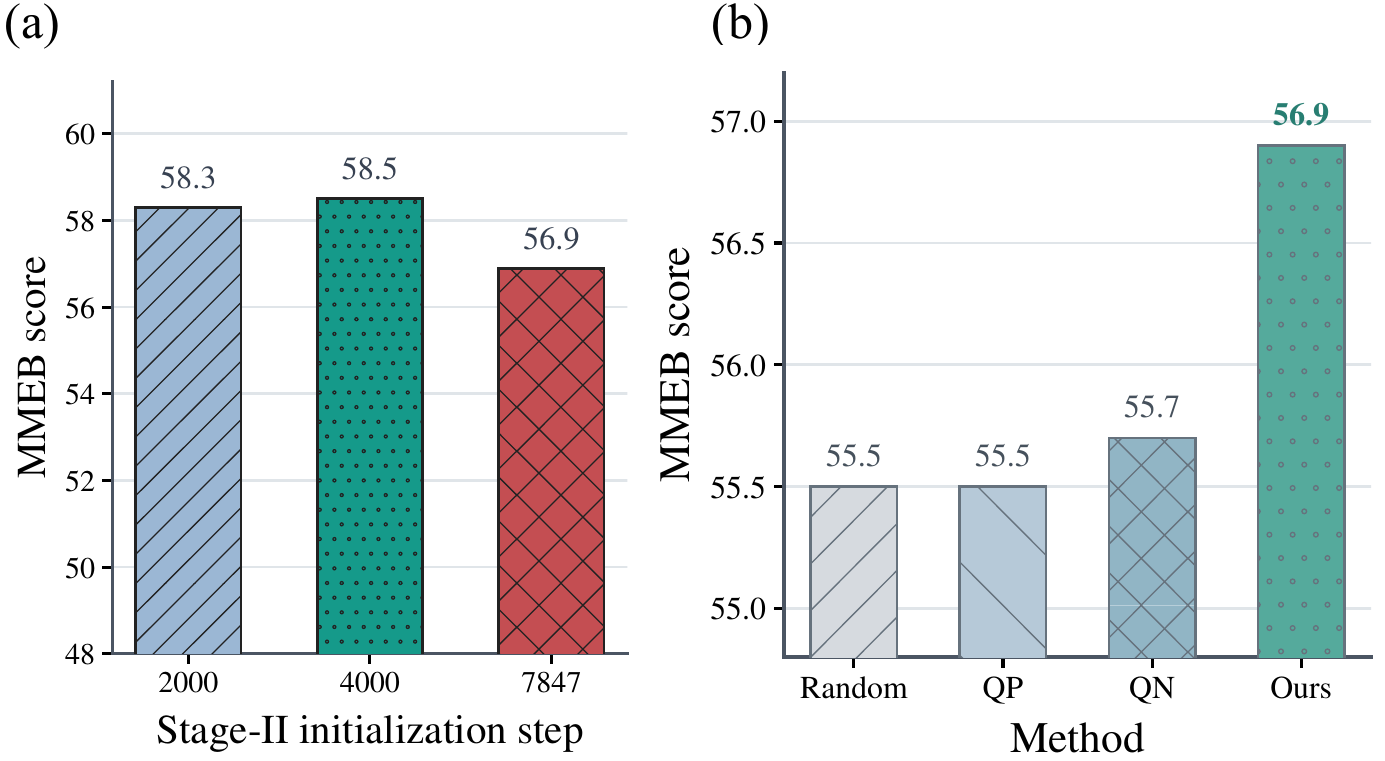}
  \caption{Ablations of Stage~I design choices. (a)~Intermediate 
  checkpoints provide better initialization for Stage~II than the final. 
  (b)~Focal Embedding Loss outperforms random, query-positive~(QP) and query-negative~(QN) weighting.}
  \label{fig:abla_step_and_focal}
\end{figure}

\noindent\textbf{Stage-Transition Checkpoint.}
Figure~\ref{fig:abla_step_and_focal} (a) examines the effect of the Stage~II initialization checkpoint. 
Initializing Stage~II from the intermediate Stage~I at steps~2000 
and 4000 yields comparable scores of 58.3 and 58.5, both exceeding the 56.9 obtained from the fully converged checkpoint at step~7847
This suggests that fully converging in the constrained first-stage
space can overspecialize the learned angular geometry and reduce plasticity when the optimization space is subsequently expanded. Intermediate checkpoints instead provide a more favorable balance between reliable semantic initialization and adaptability for Stage~II. Further analyzes of Stage~II initialization checkpoints are provided in the supplementary material. 

\noindent\textbf{Focal Embedding Loss Variants.}
To assess whether the improvement arises from difficulty-aware weighting
rather than arbitrary loss rescaling, Figure~\ref{fig:abla_step_and_focal}(b)
compares Focal Embedding Loss with three alternatives: random sample
weighting, query--positive similarity weighting, and query--negative
similarity weighting based on the hardest negative. For simplicity, all
variants are evaluated in Stage~I under otherwise identical settings. Focal
Embedding Loss performs best, indicating that query-level retrieval
confidence over the complete candidate set---which jointly reflects positive
alignment and negative competition---provides a more informative difficulty
estimate than either signal alone. Sensitivity to the focal parameter
$\gamma$ and additional ablations are provided in the supplementary material.

\section{Conclusion}
\label{sec:conclusion}
In this work, we introduced PACE, a two-stage framework for stable and
expressive multimodal embedding learning. PACE progressively expands both the
representation and parameter spaces, transitioning from cosine similarity to
dot-product similarity and from LoRA to full-parameter fine-tuning. This
coarse-to-fine strategy first establishes a reliable angular semantic geometry
before introducing radial and parameter degrees of freedom. Focal Embedding
Loss further directs optimization toward ambiguous queries while reducing
redundant updates on well-separated examples. Geometric analyses demonstrate
that PACE alleviates angular--norm coupling and directional anisotropy, while
experiments across multiple backbone scales and diverse multimodal embedding
tasks consistently validate its effectiveness. These results establish
progressive optimization-space expansion as an effective approach to better
exploiting the representational capacity of MLLM-based embedders.

\bibliography{aaai2027}

\clearpage
\appendix
\numberwithin{equation}{section}
\section{Detail Experiment Setting}
\subsection{Training Details}
\label{sec:training_details}
We provide the full training configurations of PACE in
Table~\ref{tab:training_hyperparams}, with both backbones,
Qwen3.5-0.8B-Base and Qwen3.5-2B-Base, trained under identical
hyperparameter settings. In the first stage, we perform
parameter-efficient fine-tuning with LoRA (rank=32) under the Focal
Embedding Loss with global weight normalization ($\gamma=2.0$). In the
second stage, the model is initialized from the step-2,000 checkpoint of
the first stage and undergoes full-parameter fine-tuning of the language
model with dot-product similarity, with the learning rate reduced to
$2\times10^{-6}$. In both stages, the vision encoder and multimodal
projector are frozen, and each stage is trained for one epoch on
8$\times$NVIDIA H100 (80GB) GPUs using bfloat16 precision.

\begin{table}[t]
\small
\centering
\begin{tabular}{lcc}
\toprule
\textbf{Hyperparameter} & \textbf{Stage 1} & \textbf{Stage 2} \\
\midrule
Training data & \multicolumn{2}{c}{2.0M} \\
Training epochs & \multicolumn{2}{c}{1} \\
Batch size & \multicolumn{2}{c}{32} \\ 
Learning rate & $2\times10^{-5}$ & $2\times10^{-6}$ \\
Warmup ratio & \multicolumn{2}{c}{0.05} \\
Temperature &  0.1 & 100 \\
Trainable parameters & LoRA (rank=32) & Full LM \\
Initialization & Backbone & Stage~I step~2,000 \\
Loss function & \multicolumn{2}{c}{Focal Embedding Loss ($\gamma=2.0$)} \\
Similarity metric & Cosine & Dot product \\ 
Frozen modules & \multicolumn{2}{c}{Vision encoder, projector} \\
\#Hard negatives & \multicolumn{2}{c}{5} \\
Max sequence length & \multicolumn{2}{c}{8,192} \\
Max image pixels & \multicolumn{2}{c}{1,605,632} \\
Optimizer & \multicolumn{2}{c}{AdamW} \\ 
Precision & \multicolumn{2}{c}{BF16} \\
DeepSpeed Stage & \multicolumn{2}{c}{2} \\
GPU configuration & \multicolumn{2}{c}{8$\times$NVIDIA H100 (80GB)} \\
\bottomrule
\end{tabular}
\caption{Training hyperparameters and computational requirements for PACE
(Qwen3.5-0.8B-Base and Qwen3.5-2B-Base).}
\label{tab:training_hyperparams}
\end{table}

For the two fine-tuned baselines, VLM2Vec~\citep{jiang2024vlm2vec} and
UniME-V2~\citep{gu2026unime}, both of which are cosine-similarity-based,
we adopt the training configuration of our first stage, with both
baselines trained under hyperparameter settings identical to those in
Table~\ref{tab:training_hyperparams}. Their full training configurations
are summarized in Table~\ref{tab:baseline_hyperparams}. GradCache
is employed in neither PACE nor the two baselines; nevertheless,
additional experiments presented later show that GradCache can be
effectively applied to our approach.

\begin{table}[t]
\small
\centering
\begin{tabular}{lcc}
\toprule
\textbf{Hyperparameter} & \textbf{VLM2Vec} & \textbf{UniME-V2} \\
\midrule
Training data & \multicolumn{2}{c}{2.0M} \\
Training epochs & \multicolumn{2}{c}{1} \\
Batch size & \multicolumn{2}{c}{32} \\
Learning rate & \multicolumn{2}{c}{$2\times10^{-5}$} \\
Warmup ratio & \multicolumn{2}{c}{0.05} \\
Temperature & \multicolumn{2}{c}{0.1} \\
Trainable parameters & \multicolumn{2}{c}{LoRA (rank=32)} \\
Initialization & \multicolumn{2}{c}{Backbone} \\
Loss function & InfoNCE & MLLM-judge InfoNCE \\ 
Similarity metric & \multicolumn{2}{c}{Cosine} \\
Frozen modules & \multicolumn{2}{c}{Vision encoder, projector} \\
\#Hard negatives & \multicolumn{2}{c}{5} \\
Max sequence length & \multicolumn{2}{c}{8,192} \\
Max image pixels & \multicolumn{2}{c}{1,605,632} \\
Optimizer & \multicolumn{2}{c}{AdamW} \\
Precision & \multicolumn{2}{c}{BF16} \\
DeepSpeed Stage & \multicolumn{2}{c}{2} \\
GPU configuration & \multicolumn{2}{c}{8$\times$NVIDIA H100 (80GB)} \\
\bottomrule
\end{tabular}
\caption{Training hyperparameters and computational requirements for the
retrained baselines VLM2Vec and UniME-V2.}
\label{tab:baseline_hyperparams}
\end{table}

\subsection{Instruction Prompts}
\label{sec:instruction_prompts}

Every query and candidate is wrapped with the unified
\texttt{qwen3\_5\_emb} chat template shown below, which injects the
system-level instruction ``Represent the user's input.'' automatically.

Each query, positive, and hard negative is encoded independently under
this template, and the hidden state at the last \texttt{<|endoftext|>}
token is used for last-token pooling to obtain the embedding. Images and
videos are inserted into the user content through \texttt{<image>} and
\texttt{<video>} placeholders together with their corresponding media
inputs.
\begin{quote}
\begin{verbatim}
<|im_start|>system
Represent the user's input.
<|im_end|>
<|im_start|>user
{content}
<|im_end|>
<|im_start|>assistant
<|endoftext|>
\end{verbatim}
\end{quote}

Task-level instructions are written directly into the user content of
the query; Table~\ref{tab:instruction_patterns} lists the instructions
used for each task format. Unlike reranking-style prompting, which
concatenates a query and a candidate into a single Yes/No judgment
prompt, the embedding model encodes every instance separately and
computes similarities between the resulting embeddings. All PACE models
and retrained baselines share this unified template and instruction set.

\begin{table*}[t]
\small
\centering
\begin{tabular}{ll}
\toprule
\textbf{Task format} & \textbf{Task-level instruction} \\
\midrule
Visual question answering & Represent the given image to answer the following question. \\
Classification & Represent the given image to answer the following question. \\ 
Text$\to$image retrieval & Retrieve an image that best matches the following description. \\
Image$\to$text retrieval & Find the caption that best describes this image. \\
Composed retrieval & Find an image matching the described changes. \\
\bottomrule
\end{tabular}
\caption{Task-level instructions written into the user content of the
query for each task format. The system-level instruction is injected
automatically by the template.}
\label{tab:instruction_patterns}
\end{table*}

\section{Detailed Theoretical Analysis}
\label{app:detailed-theory}

We provide the proof of Proposition~1. Under the common-norm setting, write
$\mathbf{q}=r\widehat{\mathbf{q}}$ and
$\mathbf{c}_j=R\widehat{\mathbf{c}}_j$, where both directional vectors have
unit norm, and let
$u_j=\widehat{\mathbf{q}}^\top\widehat{\mathbf{c}}_j$. We treat the candidate
embeddings as fixed. For notational simplicity, we omit the Stage~II
superscript and write $p_j$ for $p_j^{(2)}$ throughout this section. We further
define $\bar{u}=\sum_jp_ju_j$; all probabilities and derivatives below are
evaluated at the Stage~II initialization after differentiation. The
radial--angular gradient-component map
$\mathcal{G}(r,\widehat{\mathbf{q}})=(a_d,\mathbf{b}_d)$ has the block Jacobian
\begin{equation}
D\mathcal{G}
=
\begin{bmatrix}
\partial a_d/\partial r & D_{\widehat{\mathbf{q}}}a_d\\
\partial\mathbf{b}_d/\partial r & D_{\widehat{\mathbf{q}}}\mathbf{b}_d
\end{bmatrix}.
\label{eq:gradient-component-jacobian}
\end{equation}
The proposition bounds its two off-diagonal blocks at the Stage~II
initialization.

\noindent\textbf{Proof.}
Because
$p_j=\exp(rRu_j/\tau_2)/\sum_k\exp(rRu_k/\tau_2)$, differentiating with
respect to $r$ while holding $\widehat{\mathbf{q}}$ fixed gives
\begin{equation}
\frac{\partial p_j}{\partial r}
=\frac{R}{\tau_2}p_j(u_j-\bar{u}).
\label{eq:posterior-radial-derivative}
\end{equation}
Since $\widehat{\mathbf{q}}$ and the candidate embeddings are held fixed,
$\mathbf{P}_{\widehat{\mathbf{q}}}^{\perp}$ and
$\widehat{\mathbf{c}}_j$ are independent of $r$. Therefore,
\begin{equation}
\begin{aligned}
\frac{\partial\mathbf{b}_d}{\partial r}
&=
\frac{R}{\tau_2}
\mathbf{P}_{\widehat{\mathbf{q}}}^{\perp}
\sum_j\frac{\partial p_j}{\partial r}
\widehat{\mathbf{c}}_j\\
&=
\frac{R^2}{\tau_2^2}
\mathbf{P}_{\widehat{\mathbf{q}}}^{\perp}
\sum_jp_j(u_j-\bar{u})\widehat{\mathbf{c}}_j.
\end{aligned}
\label{eq:tangential-gradient-radial-derivative}
\end{equation}
Since $\sum_jp_j(u_j-\bar{u})=0$, the centered sum can be rewritten as
\begin{equation}
\sum_jp_j(u_j-\bar{u})\widehat{\mathbf{c}}_j
=
\sum_{j\neq+}p_j(u_j-\bar{u})
(\widehat{\mathbf{c}}_j-\widehat{\mathbf{c}}_+).
\label{eq:centered-candidate-sum}
\end{equation}
The unit-norm assumptions imply
$|u_j-\bar{u}|\leq2$ and
$\|\widehat{\mathbf{c}}_j-\widehat{\mathbf{c}}_+\|_2\leq2$.
Moreover, $p_+\geq1-\rho$ gives
$\sum_{j\neq+}p_j\leq\rho$. Because orthogonal projection is
nonexpansive,
\begin{equation}
\left\|\frac{\partial\mathbf{b}_d}{\partial r}\right\|_2
\leq
\frac{4R^2}{\tau_2^2}\rho.
\label{eq:radial-to-angular-bound}
\end{equation}

For the reverse coupling, let $\mathbf{h}\perp\widehat{\mathbf{q}}$ with
$\|\mathbf{h}\|_2=1$, and define
$v_j=\mathbf{h}^{\top}\widehat{\mathbf{c}}_j$ and
$\bar{v}=\sum_jp_jv_j$. Along a smooth curve on the unit sphere with tangent
$\mathbf{h}$, $D_{\widehat{\mathbf{q}}}u_j[\mathbf{h}]=v_j$. Holding
$r=r_0$ fixed, define the Stage~II logit $z_j=\beta u_j$, where
$\beta=r_0R/\tau_2$. Since
$D_{\widehat{\mathbf{q}}}z_j[\mathbf{h}]=\beta v_j$, the softmax
differential gives
\begin{equation}
\begin{aligned}
D_{\widehat{\mathbf{q}}}p_j[\mathbf{h}]
&=p_j\left(\beta v_j-\beta\sum_kp_kv_k\right)\\
&=\beta p_j(v_j-\bar{v}).
\end{aligned}
\label{eq:posterior-angular-derivative}
\end{equation}
Applying the product rule to
$a_d=(R/\tau_2)(\sum_jp_ju_j-u_+)$ gives
\begin{equation}
\begin{aligned}
D_{\widehat{\mathbf{q}}}a_d[\mathbf{h}]
&=\frac{R}{\tau_2}\bigg[
\sum_jD_{\widehat{\mathbf{q}}}p_j[\mathbf{h}]u_j\\
&\quad+\sum_jp_jD_{\widehat{\mathbf{q}}}u_j[\mathbf{h}]
-D_{\widehat{\mathbf{q}}}u_+[\mathbf{h}]\bigg]\\
&=\frac{R}{\tau_2}\bigg[
\beta\sum_jp_j(v_j-\bar{v})u_j
+\sum_jp_jv_j-v_+\bigg].
\end{aligned}
\label{eq:radial-coefficient-product-rule}
\end{equation}
Since
$\sum_jp_j(v_j-\bar{v})u_j
=\mathbb{E}_{p}[uv]-\mathbb{E}_{p}[u]\mathbb{E}_{p}[v]
=\operatorname{Cov}_{p}(u,v)$ and
$\sum_jp_jv_j=\mathbb{E}_{p}[v]$, this becomes
\begin{equation}
D_{\widehat{\mathbf{q}}}a_d[\mathbf{h}]
=
\frac{R}{\tau_2}
\left[
\beta\operatorname{Cov}_{p}(u,v)
+\mathbb{E}_{p}[v]-v_+
\right].
\label{eq:exact-angular-to-radial-derivative}
\end{equation}
Under $p_+\geq1-\rho$,
\begin{equation}
\begin{aligned}
\operatorname{Var}_{p}(u)
&\leq \sum_{j\neq+}p_j(u_j-u_+)^2
\leq 4\rho,\\
\operatorname{Var}_{p}(v)
&\leq \sum_{j\neq+}p_j(v_j-v_+)^2
\leq 4\rho.
\end{aligned}
\label{eq:variance-bounds}
\end{equation}
By Equation~\ref{eq:variance-bounds} and the Cauchy--Schwarz inequality,
$|\operatorname{Cov}_{p}(u,v)|\leq4\rho$. In addition,
\begin{equation}
\left|\mathbb{E}_{p}[v]-v_+\right|
=
\left|\sum_{j\neq+}p_j(v_j-v_+)\right|
\leq2\rho.
\label{eq:mean-displacement-bound}
\end{equation}
Consequently, for every unit tangent direction $\mathbf{h}$,
\begin{equation}
\left|D_{\widehat{\mathbf{q}}}a_d[\mathbf{h}]\right|
\leq
\frac{R}{\tau_2}(4\beta+2)\rho.
\label{eq:directional-radial-bound}
\end{equation}
Taking the supremum over such $\mathbf{h}$ proves
\begin{equation}
\left\|D_{\widehat{\mathbf{q}}}a_d\right\|_{\mathrm{op}}
\leq
\frac{R}{\tau_2}(4\beta+2)\rho,
\label{eq:angular-to-radial-operator-bound}
\end{equation}
which completes the proof. 

The result controls the local off-diagonal Jacobian blocks of the
radial--angular gradient-component map under the common-norm assumption. It
does not imply global statistical independence or a global convergence
guarantee.

\section{Additional Experiments}

\subsection{Effect of GradCache}

\begin{table}[t]
    \centering
    \setlength{\tabcolsep}{5.0pt}

    \begin{tabular}{l c c c c c}
        \toprule
        \textbf{Method}
        & \textbf{CLS}
        & \textbf{VQA}
        & \textbf{RET}
        & \textbf{GRD}
        & \textbf{Overall} \\
        \midrule

        VLM2Vec
        & 51.0 & 56.8 & 53.1 & 73.2 & 55.8\\

        VLM2Vec$^{\ddagger}$
        & 50.7 & 61.6 & 55.6 & 73.8 & 57.9\\

        PACE
        & 52.3 & 60.4 & 57.4 & 77.2 & \underline{59.0}\\

        PACE$^{\ddagger}$
        & 53.7 & 63.7 & 58.4 & 74.5 & \textbf{60.4}\\

        \bottomrule
    \end{tabular}
    \caption{%
        Effect of GradCache on MMEB with Qwen3.5-0.8B-Base.
        $^{\ddagger}$ denotes training with GradCache enabled.
        CLS: Classification, RET: Retrieval, GRD: Grounding.
        GradCache improves the overall score of both methods, with PACE
        retaining its lead. Best results are in bold and second-best are
        underlined.
    }
    \label{tab:gradcache}
\end{table}

\noindent\textbf{Effect of GradCache.}
As noted in the training details, GradCache is employed in neither PACE nor the retrained baselines.
To verify its compatibility with our approach, we retrain PACE and VLM2Vec with GradCache enabled, with all hyperparameter settings kept unchanged except the temperature, which is reduced to one fifth of its original value, and report the per-category results in Table~\ref{tab:gradcache}.

GradCache lifts the overall score of both methods---from 59.0 to 60.4 for PACE and from 55.8 to 57.9 for VLM2Vec---with gains concentrated in VQA and retrieval, where the enlarged effective batch supplies more informative in-batch negatives.
PACE with GradCache retains a 2.5-point overall lead over its VLM2Vec counterpart, which indicates that the benefit of the staged training is complementary to memory-efficient large-batch optimization rather than subsumed by it.
The only regression is the grounding score of PACE (77.2 to 74.5), which nevertheless remains the highest among all configurations.

\subsection{Additional Geometry Analysis}
\begin{figure*}[!t]
  \centering
  \includegraphics[width=\textwidth]{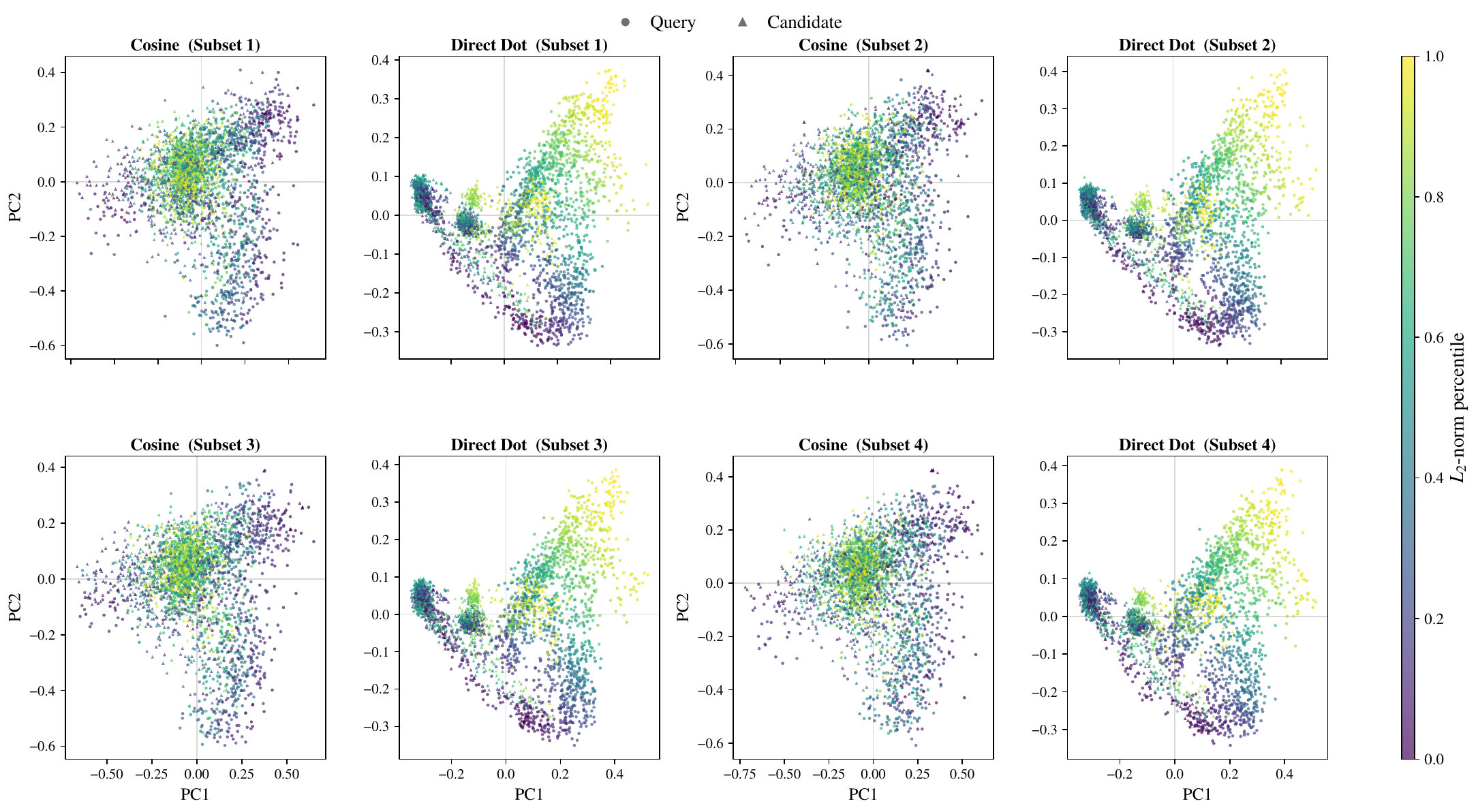}
  \caption{PCA visualizations of $\ell_2$-normalized embeddings on four
randomly resampled subsets, colored by the $L_2$-norm percentile of the
original embeddings; circles and triangles denote queries and candidates.
Across all subsets, cosine-based training yields intermixed colors,
indicating low mutual information between directions and norms, while
direct dot-product training reproduces the same norm-stratified layout,
confirming that the entanglement pattern in Figure~\ref{fig:motivation}(a)
is robust to sampling.}
  \label{fig:appendix_pca_resamples}
\end{figure*}

\noindent\textbf{PCA Visualization across Resampled Subsets.}
To verify that the qualitative pattern in Figure~\ref{fig:motivation}(a) is not an artifact of a particular sample, Figure~\ref{fig:appendix_pca_resamples} repeats the analysis on four subsets independently drawn from the evaluation data with different random seeds, applying the same procedure: embeddings are $\ell_2$-normalized, projected by PCA, and colored by their original norm percentile.

The observation is consistent across all four subsets.
Under cosine-based training, embeddings with different norms remain intermixed in the projected space, and no norm-dependent structure emerges in any subset---directions carry little information about norms, implying low mutual information between the two factors.
Under direct dot-product training, every subset reproduces the same norm-stratified layout, with high-norm embeddings concentrated in one projected region and low-norm embeddings in another.
Moreover, the norm percentile varies smoothly along a coherent low-dimensional structure in the projected directional space, so that embedding magnitude is largely predictable from direction alone---a direct signature of the coupling between norms and angles.
This stability across resampled subsets confirms that the angular--norm entanglement induced by direct dot-product optimization is a systematic property of the learned embedding space rather than a sampling artifact.

\begin{figure}[!t]
  \centering
  \includegraphics[width=0.7\columnwidth]{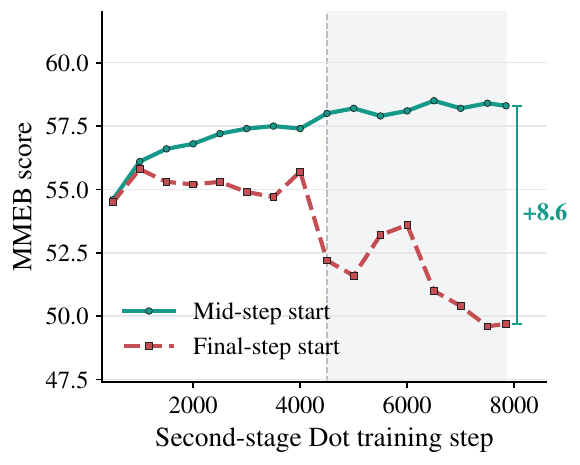}
  \caption{Stage~II training dynamics under different Stage~I
initializations. MMEB score is tracked along the second-stage
dot-product training trajectory, with \emph{Mid-step start} initialized
from the Stage~I checkpoint at step~4{,}000 and \emph{Final-step start}
from the fully converged checkpoint at step~7{,}847. Mid-step start
converges stably to 58.3, while Final-step start deteriorates after
step~4{,}500 (shaded region), leaving an 8.6-point gap at convergence.}
  \label{fig:appendix_midstep_vs_final}
\end{figure}

\noindent\textbf{Staged Training Alleviates the Coupling.}
\begin{figure}[!t]
  \centering
  \includegraphics[width=\columnwidth]{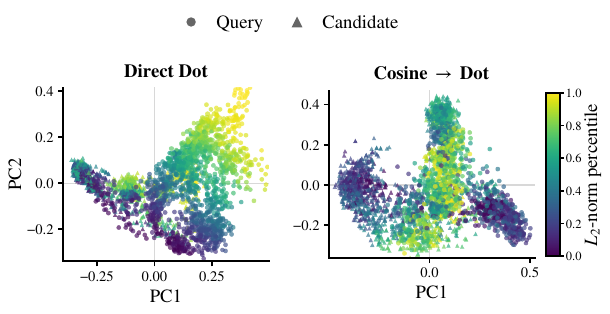}
  \caption{PCA visualizations of $\ell_2$-normalized embeddings under
direct dot-product training (left) and staged cosine$\to$dot
training (right), colored by $L_2$-norm percentile. Direct
dot-product training exhibits the norm-stratified low-dimensional
structure analyzed above, while staged training restores intermixed
colors---norms are no longer predictable from directions---indicating
reduced mutual information between the two factors.}
  \label{fig:appendix_pca_d2c}
\end{figure}

Figure~\ref{fig:appendix_pca_d2c} compares direct dot-product training with the staged cosine$\to$dot schedule under the same visualization protocol.
As in the preceding experiments, focal weighting is disabled, isolating the effect of the stage schedule itself.
Direct dot-product training reproduces the coherent low-dimensional structure identified above, with the norm percentile varying smoothly along the projected directions.
Under the staged schedule, this structure disappears: embeddings of different norms are intermixed across the projected space, and no norm-aligned gradient remains.
The cosine$\to$dot transition alone therefore suffices to alleviate the angular--norm coupling induced by direct dot-product optimization, letting the model encode magnitude as a complementary signal rather than binding it to particular directions.

\subsection{Analysis of Stage Transition}

\noindent\textbf{Stage-Transition Training Dynamics.}
To complement the summary in the main text, Figure~\ref{fig:appendix_midstep_vs_final} traces the full Stage~II training trajectory for two representative initializations: \emph{Mid-step start} (from the Stage~I checkpoint at step~4{,}000) and \emph{Final-step start} (from the fully converged checkpoint at step~7{,}847).
As in the corresponding main-text ablation, this experiment disables focal weighting; the Stage~I checkpoint steps therefore differ from the step-2{,}000 initialization used in the full PACE configuration.

Mid-step start yields a smooth, monotonically increasing MMEB curve that plateaus around 58.3 after approximately 4{,}500 second-stage steps.
The Final-step start, by contrary, peaks early at 55.8 within the first 1{,}000 steps, oscillates around 55 thereafter, and deteriorates with pronounced fluctuations after step~4{,}500, ultimately falling to 49.7, an 8.6-point gap relative to the Mid-step start at convergence.

This divergence corroborates the hypothesis of overspecialization outlined in the main text.
Stage~I optimizes a cosine-based InfoNCE loss that projects embeddings onto the unit hypersphere, learning angular alignment while discarding norm information.
A fully converged Stage~I checkpoint has thoroughly committed to this norm-invariant geometry.
When the objective switches to dot-product similarity in Stage~II---requiring the model to additionally leverage embedding magnitude as a relevance signal---the optimization must simultaneously restructure the angular layout and recover norm sensitivity.
The conflicting gradient signals destabilize training and lead to the degradation observed after step~4{,}500.
In contrast, an intermediate Stage~I checkpoint has acquired sufficient directional structure to provide a reliable semantic initialization, yet retains enough representational plasticity for the model to incorporate magnitude information smoothly under the expanded optimization space of Stage~II.

\subsection{Sensitivity to the Focal Parameter}
\label{sec:gamma-sensitivity}

\begin{table}[t]
    \centering
    \setlength{\tabcolsep}{10pt}
    \begin{tabular}{c cc}
        \toprule
        \textbf{$\gamma$} & \textbf{After Stage~I} & \textbf{After Stage~II} \\
        \midrule
        0.5 & 56.3 & 58.2 \\
        2.0 & \textbf{56.9} & \textbf{59.0} \\
        4.0 & \textbf{56.9} & 58.7 \\
        \bottomrule
    \end{tabular}
    \caption{Sensitivity to the focal parameter $\gamma$ on MMEB using
    Qwen3.5-0.8B-Base. Scores denote average Precision@1 (\%) after each training
    stage; the best result in each column is highlighted in bold.}
    \label{tab:gamma-sensitivity}
\end{table}

Table~\ref{tab:gamma-sensitivity} reports the sensitivity of PACE to the focal
parameter $\gamma$ under otherwise identical training settings. Stage~I
performance varies by only 0.6 across the evaluated values. Stage~II
improves every configuration by 1.8--2.1, yielding final scores between
58.2 and 59.0. These results indicate that PACE exhibits limited sensitivity to
$\gamma$ within the evaluated range. The setting $\gamma=2.0$ achieves the
highest final score of 59.0, while stronger modulation provides no additional
benefit. We therefore adopt $\gamma=2.0$ as the default.

\section{External Results}
Table~\ref{tab:mmeb_detailed} reports the per-dataset results of PACE
and all compared methods on the MMEB benchmark.
\begin{table*}[t]

\centering
\setlength{\tabcolsep}{5.5pt}
\begin{tabular}{lcccccc}
\toprule
  & VLM2Vec &VLM2Vec$^{\dagger}$ & UniME-V2 & UniME-V2$^{\dagger}$ & PACE & PACE$^{\dagger}$ \\
\midrule
\multicolumn{7}{l}{\textbf{Classification (10 tasks)}} \\
ImageNet-1K & 55.6 & 72.6 & 58.2 & 72.0 & 59.6 & 72.6 \\
N24News & 65.7 & 69.8 & 59.4 & 59.3 & 72.0 & 75.5 \\
HatefulMemes & 49.9 & 57.3 & 45.8 & 43.2 & 49.1 & 58.4 \\
VOC2007 & 74.4 & 71.1 & 81.3 & 83.4 & 77.3 & 83.6 \\
SUN397 & 67.4 & 73.0 & 67.2 & 74.9 & 65.5 & 77.5 \\
Place365$^{\ast}$ & 35.0 & 42.3 & 34.7 & 41.0 & 35.0 & 40.3 \\
ImageNet-A$^{\ast}$ & 20.9 & 49.0 & 22.5 & 51.2 & 21.8 & 50.0 \\
ImageNet-R$^{\ast}$ & 72.9 & 88.6 & 75.6 & 87.4 & 76.0 & 88.5 \\
ObjectNet$^{\ast}$ & 61.6 & 75.3 & 63.4 & 77.9 & 59.0 & 74.6 \\
Country-211$^{\ast}$ & 6.8 & 14.6 & 8.7 & 17.0 & 7.6 & 15.2 \\
All Classification & 51.0 & 61.4 & 51.7 & 60.7 & 52.3 & 63.6 \\
\midrule
\multicolumn{7}{l}{\textbf{VQA (10 tasks)}} \\
OK-VQA & 47.2 & 56.7 & 48.7 & 54.2 & 51.2 & 60.9 \\
A-OKVQA & 41.0 & 48.9 & 43.9 & 49.6 & 44.3 & 52.7 \\
DocVQA & 91.6 & 92.4 & 93.5 & 94.4 & 94.0 & 95.2 \\
InfographicsVQA & 67.6 & 72.6 & 66.1 & 71.5 & 70.4 & 75.7 \\
ChartQA & 58.9 & 63.5 & 55.0 & 61.8 & 61.2 & 64.4 \\
Visual7W & 55.0 & 60.6 & 53.9 & 57.7 & 59.6 & 63.8 \\
ScienceQA & 29.9 & 40.5 & 31.9 & 45.6 & 29.2 & 43.7 \\
VizWiz$^{\ast}$ & 50.0 & 53.6 & 51.1 & 55.3 & 51.0 & 54.9 \\
GQA$^{\ast}$ & 52.3 & 48.0 & 65.1 & 65.4 & 62.0 & 62.1 \\
TextVQA$^{\ast}$ & 74.6 & 82.9 & 81.1 & 85.5 & 81.4 & 86.3 \\
All VQA & 56.8 & 62.0 & 59.0 & 64.1 & 60.4 & 66.0 \\
\midrule
\multicolumn{7}{l}{\textbf{Retrieval (12 tasks)}} \\
VisDial & 68.3 & 71.5 & 71.6 & 77.3 & 77.9 & 81.8 \\
CIRR & 42.6 & 55.5 & 32.1 & 49.3 & 41.8 & 51.6 \\
VisualNews$_{\mathrm{t2i}}$ & 46.8 & 61.4 & 50.9 & 64.5 & 50.8 & 65.4 \\
VisualNews$_{\mathrm{i2t}}$ & 50.4 & 64.2 & 54.8 & 67.5 & 53.9 & 70.0 \\
MSCOCO$_{\mathrm{t2i}}$ & 66.5 & 73.2 & 70.8 & 76.2 & 71.7 & 76.5 \\
MSCOCO$_{\mathrm{i2t}}$ & 65.0 & 67.4 & 66.4 & 70.0 & 69.8 & 72.8 \\
NIGHTS & 64.7 & 65.3 & 62.5 & 66.6 & 65.1 & 67.1 \\
WebQA & 82.0 & 85.7 & 74.0 & 82.0 & 78.2 & 88.2 \\
FashionIQ$^{\ast}$ & 16.9 & 18.7 & 9.8 & 22.5 & 19.5 & 21.1 \\
Wiki-SS-NQ$^{\ast}$ & 58.6 & 63.3 & 58.1 & 66.5 & 61.3 & 66.8 \\
OVEN$^{\ast}$ & 35.0 & 46.4 & 45.6 & 62.7 & 44.2 & 65.7 \\
EDIS$^{\ast}$ & 40.6 & 64.0 & 49.2 & 60.3 & 54.5 & 71.5 \\
All Retrieval & 53.1 & 61.4 & 53.8 & 63.8 & 57.4 & 66.5 \\
\midrule
\multicolumn{7}{l}{\textbf{Visual Grounding (4 tasks)}} \\
MSCOCO$^{\ast}$ & 60.8 & 60.3 & 62.1 & 65.1 & 63.2 & 69.7 \\
RefCOCO$^{\ast}$ & 81.1 & 85.0 & 84.2 & 89.5 & 86.5 & 91.7 \\
RefCOCO-matching$^{\ast}$ & 83.4 & 80.6 & 88.4 & 87.6 & 80.8 & 85.1 \\
Visual7W-pointing$^{\ast}$ & 67.7 & 68.9 & 65.4 & 78.9 & 78.2 & 80.8 \\
All Visual Grounding & 73.2 & 73.7 & 75.0 & 80.3 & 77.2 & 81.8 \\
\midrule
\multicolumn{7}{l}{\textbf{Final Score (36 tasks)}} \\
All IND & 59.5 & 66.2 & 59.4 & 66.0 & 62.1 & 69.9 \\
All OOD & 51.1 & 58.8 & 54.1 & 63.4 & 55.1 & 64.0 \\
All & 55.8 & 62.9 & 57.0 & 64.9 & 59.0 & 67.3 \\
\bottomrule
\end{tabular}
\caption{Per-dataset results on the MMEB benchmark, with ``All'' denoting the
average over all 36 datasets. Datasets marked with $^{\ast}$ denote OOD.
$^{\dagger}$ indicates the Qwen3.5-2B-Base backbone; all other methods
use Qwen3.5-0.8B-Base.}
\label{tab:mmeb_detailed}
\end{table*}
\end{document}